\RequirePackage{etex} %% **************** add this for ARXIV submission
\documentclass[runningheads,twocolumn,a4paper,10pt]{llncs}
\usepackage{etex}
\reserveinserts{28}
\pdfoutput=1
\usepackage[a4paper, total={7in, 10in}]{geometry}
\usepackage{ textcomp }
\usepackage{ comment }
\usepackage[color=yellow!60,textsize=footnotesize,obeyDraft,draft]{todonotes}
\usepackage{colortbl}
\usepackage{booktabs}
\usepackage{capt-of}
\usepackage{makecell}
\usepackage{multirow}
\usepackage{listings}
\usepackage{graphicx}
\usepackage{amsmath}
\usepackage{caption}
\usepackage{subcaption}
\usepackage[font=footnotesize,labelfont=bf]{subcaption}
\usepackage[hidelinks]{hyperref}
\usepackage[capitalize]{cleveref}
\usepackage{tikz}
\usepackage{pgfplots}
\pgfplotsset{compat=newest}
\pgfplotsset{plot coordinates/math parser=false}
\usepackage[para,online,flushleft]{threeparttable}
\usepackage{wasysym}
\usepackage{amssymb}
\usetikzlibrary{arrows}
\usepackage{breakcites}
\usepackage{array}
\usepackage{color}
\usepackage{balance} %balance columns
\usepackage{lastpage}
\usepackage{dblfloatfix}
\usepackage{authblk}
\usepackage{siunitx}
\usepackage{xurl} %for long URLs
\usepackage[shortlabels]{enumitem}
\usepackage{soul}
\usepackage{xcolor,listings}
\usepackage{relsize}
\usepackage{algorithm}
\usepackage{algpseudocode}

\usepackage{float}
\floatstyle{plaintop}
\restylefloat{table}
\usepackage{tikz}
\usetikzlibrary{shapes.geometric, arrows,calc}
\tikzstyle{startstop} = [rectangle, rounded corners, minimum width=3cm, minimum height=1cm,text centered, draw=black, fill=none]
\tikzstyle{io} = [trapezium, trapezium left angle=70, trapezium right angle=110, minimum width=3cm, minimum height=1cm, text centered, draw=black, fill=blue!30]
\tikzstyle{process} = [rectangle, minimum width=3.5cm, minimum height=1cm, text centered, draw=black, fill=orange!0]
\tikzstyle{neuron} = [circle, minimum size=3cm, text centered, draw=black, fill=orange!0]
\tikzstyle{decision} = [diamond, minimum width=3cm, minimum height=1cm, text centered, draw=black, fill=green!30]
\tikzstyle{arrow} = [thick,->,>=stealth]

\usepackage[maxnames=6,firstinits=true,doi=false,url=true,isbn=false]{biblatex}
\begin{document} 

% \title{Towards a Measure of Trustworthiness to Evaluate CNNs During Operation}
\title{A Trustworthiness Score to Evaluate DNN Predictions}%A Measure of Trustworthiness to Evaluate CNNs Predictions}
\titlerunning{Trustworthiness Score to Evaluate DNN Predictions}
\authorrunning{Ghobrial et al.}

\author{Abanoub Ghobrial\inst{1}, Darryl Hond\inst{2}, Hamid Asgari\inst{2}, Kerstin Eder\inst{1}}

\institute{Trustworthy Systems Lab, University of Bristol, Bristol, UK \and RTI Centre, Thales, Reading, UK}
\tocauthor{Authors' Instructions}

\maketitle
\let\thefootnote\relax\footnotetext{
% \textit{Statements about authorship contribution.}
Abanoub Ghobrial (e-mail: abanoub.ghobrial@bristol.ac.uk), 
and 
Kerstin Eder (e-mail: kerstin.eder@bristol.ac.uk) 
are with the Trustworthy Systems Lab, Department of Computer Science, University of Bristol, Merchant Ventures Building, Woodland Road, Bristol, BS8 1UB, United Kingdom. 
Darryl Hond (e-mail: darryl.hond@uk.thalesgroup.com)
Hamid Asgari (e-mail: hamid.asgari@uk.thalesgroup.com) 
is with Research, Technology and Innovation (RTI) centre,Thales, Reading, United Kingdom.}

\makeatletter
\renewcommand\subsubsection{\@startsection{subsubsection}{3}{\z@}%
                       {-18\p@ \@plus -4\p@ \@minus -4\p@}%
                       {4\p@ \@plus 2\p@ \@minus 2\p@}%
                       {\normalfont\normalsize\bfseries\boldmath
                        \rightskip=\z@ \@plus 8em\pretolerance=10000 }}
\makeatother

\textbf{\textit{Abstract}--Due to the black box nature of deep neural networks (DNN), the continuous validation of DNN during operation is challenging with the absence of a human monitor. 
As a result this makes it difficult for developers and regulators to gain confidence in the deployment of autonomous systems employing DNN.
It is critical for safety during operation to know when DNN's predictions are trustworthy or suspicious.  
With the absence of a human monitor, the basic approach is to use the model's output confidence score to assess if predictions are trustworthy or suspicious.  
However, the model's confidence score is a result of computations coming from a black box, therefore lacks transparency and makes it challenging to automatedly credit trustworthiness to predictions.
We introduce the \textit{trustworthiness score} (TS), a simple metric that provides a more transparent and effective way of providing confidence in DNN predictions compared to model's confidence score. 
The metric quantifies the trustworthiness in a prediction by checking for the existence of certain features in the predictions made by the DNN. 
We also use the underlying idea of the TS metric, to provide a \textit{suspiciousness score} (SS) in the overall input frame to help in the detection of suspicious frames where false negatives exist.
% which provides for suspiciousness in a frame  by scanning for the existence of untrustworthy predictions.
%
We conduct a case study using YOLOv5 on persons detection to demonstrate our method and usage of TS and SS. 
The case study shows that using our method consistently improves the precision of predictions compared to relying on model confidence score alone, for both 1) approving of trustworthy predictions ($\sim 20\%$ improvement) and 2) detecting suspicious frames ($\sim 5\%$ improvement). }

% ***************************************************
%  Main Body
% ***************************************************

% ========================
% Introduction
% ========================
\section{Introduction}
Deep neural networks (DNN) for object detection are being  heavily adopted in the perception task of autonomous systems.
Despite the power of DNN at being able to learn complicated patterns from complex environments and producing highly non linear decision boundaries, yet they are still prone to outputting miss-classifications~\cite{Ghobrial2022}. 
Affected by the black box nature of DNN in computing predictions, it makes validation of outputted predictions during operation challenging.
% the task of validation of the classifier's outputs during operation infeasible. 
%
This may hinder the deployment of autonomous systems employing such DNN.
We propose that confidence can be gained to deploy such systems if the system continuously provides certain levels of trustworthiness in their predictions during operation.
In this paper we introduce the \textit{trustworthiness score (TS)}, a metric that attempts to provide a scalar quantification of trustworthiness in DNN predictions. 
This is achieved by monitoring and checking for overlaps of \textit{features specifications} with predictions. 
The term features specifications is used here to refer to unique distinctive properties that make up a classification class.
The TS metric can serve as an online monitoring tool to improve the evaluation of DNN trustworthy predictions during operation, whilst also providing a transparent justification for accrediting trustworthiness to predictions. 
We also explore if using the underlying idea of the trustworthiness score in a prediction can be used to calculate a suspiciousness score (SS) in the overall input frame, which may help in the detection of suspicious frames where false negatives exist.
% nd assist with miss-classifications detection. 
%
% This may serve as an assistant to the ground truth monitor (human) in monitoring, or substitute the ground truth where it is infeasible to have a human ground truth monitor.
%
In the rest of this paper: we discuss related works in section~\ref{sec:related_work}, section~\ref{sec:method} introduces our method,
sections~\ref{sec:experiment} and ~\ref{sec:results} show cases our approach and discusses results, finally we conclude in section~\ref{sec:conclusions}.

% ========================
% Related Work
% ========================
\section{Related Work}\label{sec:related_work}
%
% using explainability in machine learning and features specficiations  explanations and reasoning similar to humans reasoning of trust.Thus can be used in   and can be used to provide accountability \textit{Trust in Classification Score}
%
We discuss relevant parts of trust in humans vs trustworthiness, overview existing works relating to trust calculations in machine learning (ML), discuss why existing methods are insufficient as trustworthiness metrics, contrast our proposed approach against the notion of \textit{uncertainty quantification} in ML and lastly review relevant explainability techniques. %relevant to our proposed methodology. 
%
% \textcolor{red}{
%, often represented as a probability that a model will make the correct decision.}

\subsection{Trust and Trustworthiness}\label{trustVStrustowrthiness}
\textit{Trust} is a term that is conceptually used to express a trustor's (someone) subjective estimate of the probability that the trustee (someone or something) will display the trustor's preferred behaviour~\cite{Bauer2013}.
Therefore trust as interpreted by humans is a measure of subjective confidence and may vary and evolve dependant on an individual's personal set of skills and experiences~\cite{Hommel2015, Mitkidis2017,Najib2019}. 
% This aligns with the hypothesis that trust in humans is largely reliant on the degree of interpersonal similarities, which may .  
% Trust in humans is usually based on interpersonal similarity as discussed in \cite{Mitkidis2017}.
%
\textit{Trustworthiness} on the other hand, whilst also being a measure of confidence, is evaluated based on the trustee demonstrating a certain set of established characteristics that prove they are doing the \textit{correct} behaviour based on some ground truth reference~\cite{Bauer2013, Bournival2022}.
These demonstrable characteristics vary depending on the application and on whether the trustee being a human or a machine~\cite{Wing2021}.
Trust and trustworthiness tends to be used interchangeable in some ML literature~\cite{DeBie2021,Jiang2018}.
Following definitions discussed above we distinguish between trust and trustworthiness: trust being a subjective measure of confidence based on experiences and hence is best for usage between human to human interactions, while trustworthiness is based on demonstrating that the \textit{correct} behaviour (grounded by a set of characteristics) is pursued and thus is more suitable for runtime evaluation of ML predictions. 
Access to ground truth or a reference point is required to check for the \textit{correctness} likelihood of a prediction~\cite{Barr2015}, making the demonstration of a system's trustworthiness troublesome during runtime.

Jiang et.al.~\cite{Jiang2018} introduced the \textit{trust score}, which considers a prediction as trustworthy if the operating classifier's prediction aligns with the nearest-neighbor classifier's output based on the training data. Here the nearest-neighbor classifier based on the training data is the reference point. However, Jiang's trustworthiness calculation lack's transparency and explainability.
De Bie et.al.~\cite{DeBie2021} tried to overcome the explainability limitation with RETRO-VIZ. This method takes a very similar approach as Jiang et.al.\cite{Jiang2018} in calculating the trustworthiness using the trust score, but replaces the ground truth output of the nearest neighbour classifier with the output from a \textit{reference set}. The reference set is extracted from the training data using similarity measures~\cite{Hond2020} (e.g. euclidean distance) to select data points similar to the input data point. 
Additionally they try to improve upon Jiang et.al.'s approach by producing a visualisation that help understand the reasons for the estimated trustworthiness based on features from the reference set. This aims at providing a justification to why a prediction was assigned a certain trust score. It does not ensure the trustworthiness is calculated based on the existence of certain features or characteristics that are demonstrable in the prediction.%, and thus should make up a trustworthy prediction. 

Both approaches by Jiang and De Bie fit better under \textit{trust} calculations, which we differentiate from trustworthiness as outlined above: trustworthiness should be calculated based on certain characteristics found in a prediction rather than calculating the trustworthiness based on hidden features then try to explain which features credited trustworthiness. 
Additionally, their approaches benefit from alleviating the engineering effort needed to identify which features should matter in calculating the trustworthiness score, but as a result it may lead to misleading and unworthy trustworthiness scores if used in operation.
% by finding similarity in scores between these hidden features and other  characteristics that were used in calcuating the trustworhtiness.
%
%If the classifer's output  Thus determine whether the output of the operating classifier $h$ would be trustworthy or not. 
%
We overcome this limitation in this paper by taking an interrogative approach in assessing the operating classifier's predictions based on established characteristics/features. 
% that are chosen and approved to be monitored during operation. 
Based on these features being evident and demonstrable a trustworthiness score is assigned to the prediction.
%
% should be demonstrable in a prediction and consequently based on the evidence found we assign a trustworthiness score. 
% Due to the design nature of our approach it is transparent and thus interpretable even by non-domain experts.
% The approach we introduce in this paper for calculating the trustworthiness takes an interogative approach at calculate the trustworhtiness of a classifer's predction. Instead of  
%
% i) The methods may do not provide explanations
% Just as Both of these methods do not provide explanations to why these methods are not trustworthy, and the trustwothiness expalnations  
% This method of evaluating the similarity between input data and training data has also been used to verify or estimate the uncertainty of a classifier's output~\cite{Hond2020}.
% \cite{Najib2019}

Paudel et.al. introduced ConsensusNet, targeted at learning correlations between predictions assigned to different cells in an image by a classifier. % (e.g.YOLO~\cite{Redmon2018}). 
These correlations are hypothesised to hold for training data and operational input data. If these correlations do not hold, then the output of the classifier is considered to be a miss-classification~\cite{Paudel2021}. 
The method we introduce
% for measuring the trustworthiness of classifications 
does not use Paudel et.al.'s method of assessing correlation between grid cells
%of an image
, but we do share the underlying idea of different features that make up a label class should be monitored during operation to identify miss-classifications.

\subsection{Uncertainty in Machine Learning}
\textit{Uncertainty quantification} in ML is used to represent probabilistically the level of unknown information in a prediction, thus enabling users when to trust a model's prediction~\cite{ghahramani2015probabilistic}\cite{tran2022plex}.
One popular way for computing uncertainty is by utilising Bayesian inference in the model's architecture to output a distribution of uncertainties over outputs~\cite{gal2016dropout}\cite{ABDAR2021}. 
A drawback of that, is the need to do changes in the architecture of the neural network e.g.~\cite{Kraus2019}.
The trust score discussed earlier introduced by Jiang et.al. ~\cite{Jiang2018} overcomes the need to change the network structure and outputs only a score rather than a distribution.
Our \textit{trustworthiness score} excels beyond the aforementioned trust score in transparency, which may be very advantageous especially in human-computing interaction applications. 
Furthermore, classification with a reject option, where the model is enabled to abstain from making a prediction is very relatable to our method~\cite{hendrickx2021machine}. 
An advantage of our method, is that the model is treated as a black box, whilst classifiers with a reject option often rely on learning a classification and a rejection function simultaneously.

\subsection{Interpretability in Machine Learning}
In the context of ML, \textit{interpretability} has been very nicely defined as the ``ability to explain or to present in understandable terms to a human''~\cite{Doshi-Velez2017}. 
In turn, interpretability allows users to trust or distrust classifiers~\cite{Rudin2022}. 
Interpretability seems to be a requirement for generating trust in classifiers' predictions~\cite{Ribeiro2016, Lipton2018}, however in ~\cite{Lipton2018} it has been also argued that trust does not necessarily rely on interpretability. This latter argument can be backed up by the trust score introduced by Jiang et.al.~\cite{Jiang2018}, which does not rely on interpretability. 
However, motivated by our earlier discussion between trust and trustworthiness in section~\ref{trustVStrustowrthiness}, trustworthiness requires demonstrability therefore having interpretable \textit{explanations} becomes a requirement to generate trustworthiness in predictions.

% ========================
% Method
% ========================
\section{Method} \label{sec:method}
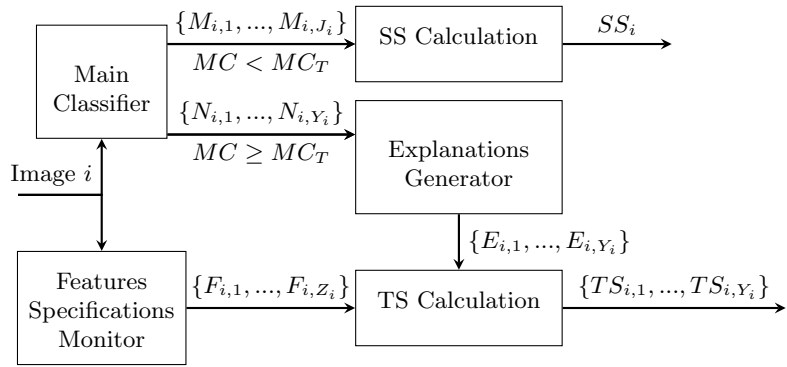
\begin{figure*}%[h]
\centering 
% \resizebox{!}{115}{
    \begin{tikzpicture}[node distance=3.5cm]
   
    \node (classifier) [process, minimum width=1.5cm, minimum height=1.5cm, text width=1.5cm, text height=0cm, text depth = 0 cm, yshift=0cm, xshift=0cm] {Main Classifier};

    % \node[text width=3cm] at ([xshift=0mm,yshift=-7mm]classifier.east) 
    % {\tiny $MC \geq MC_{T}$};

    % \node[text width=3cm] at ([xshift=0mm,yshift=5mm]classifier.east) 
    % {\tiny $MC < MC_{T}$};

    \node (ss_calculation) [process, right of=classifier, minimum width=2cm, minimum height=1cm, text width=2.5cm, text height=0cm, text depth = 0 cm, yshift=0.5cm, xshift=1.2cm] {SS Calculation};

    \node (explanations_generator) [process,  right of=classifier, minimum width=1.5cm, minimum height=1.5cm, text width=2.5cm, text height=0cm, text depth = 0 cm, yshift=-1cm, xshift=1.2cm, align=center] {Explanations Generator};
    
    \node (features) [process,  below of=classifier, minimum width=1.5cm, minimum height=1.5cm, text width=2cm, text height=0cm, text depth = 0.5 cm, yshift=0.5cm, xshift=0cm, align=center] {Features Specifications Monitor};
    
    \node (tcs_calculation) [process,  below of=explanations_generator, minimum width=2cm, minimum height=1cm, text width=2.5cm, text height=0cm, text depth = 0 cm, yshift=1.5cm, xshift=0cm, align=center] {TS Calculation};

      \draw[arrow] (-1.1,-1.5) -| node[anchor=south east] {Image ${i}$} (classifier);
      
    %   \draw (-1,0) -| node[anchor=south] {} (-1,-2.5);
      
    %   \draw[arrow] (-1,-2.5) -- node[anchor=south] {} (features);
    
    \draw[arrow] (-1.1,-1.5) -| node[anchor=south] {} (features);
      
      \draw[arrow] ([yshift=-7mm]classifier.east) -- node[anchor=south] {$\{N_{i,1},..., N_{i,Y_i}\}$} ([yshift=3mm]explanations_generator.west);
      
     \draw[arrow] ([yshift=-7mm]classifier.east) -- node[anchor=north] {$MC \geq MC_{T}$} ([yshift=3mm]explanations_generator.west);

      \draw[arrow] ([yshift=5mm]classifier.east) -- node[anchor=south] {$\{M_{i,1},..., M_{i,J_i}\}$} ([yshift=0mm]ss_calculation.west);
      
      \draw[arrow] ([yshift=5mm]classifier.east) -- node[anchor=north] {$MC < MC_{T}$} ([yshift=0mm]ss_calculation.west);

      % \draw[arrow] ([yshift=5mm]classifier.east) -- node[anchor=south] {$\{M_{i,1},..., M_{i,J_i}\}$} ([yshift=0mm]ss_calculation.west);
      
      \draw[arrow] (explanations_generator) -- node[anchor=west] {$\{E_{i,1},..., E_{i,Y_i}\}$} (tcs_calculation);
      
      \draw[arrow] (features) -- node[anchor=south] {$\{F_{i,1},..., F_{i,Z_i}\}$} (tcs_calculation);
      
      \draw[arrow] (tcs_calculation) -- node[anchor=south] {$\{TS_{i,1},..., TS_{i,Y_i}\}$} (9,-3);

      \draw[arrow] (ss_calculation) -- node[anchor=south] {$SS_{i}$} (7.5,0.5);
       
    \end{tikzpicture}
    % }
\caption{Method Overview}
\label{fig:method_overview}
\end{figure*}

An overview of the process used to calculate the trustworthiness score (TS) in predictions and the suspiciousness score in a frame is shown by 
Figure~\ref{fig:method_overview}. 
We use the term \textit{main classifier} to refer to the DNN responsible for object recognition. 
The input image is fed into the main classifier to obtain two sets of predictions. The first set of predictions $\{N_{i,1},..., N_{i,Y_i}\}$ are predictions with model confidence ($MC$) $\geq$ to the set model confidence threshold ($MC_T$). 
The second set of predictions $\{M_{i,1},..., M_{i,J_i}\}$ is for $MC < MC_T$. Where $Y_i$ and $J_i$ are the total number of predictions detected in image number $i$ for $MC \geq MC_T$ and $MC<MC_T$ respectively. 
% $i$ is the image/frame identification number.

Dependent on the type of object recognition the main classifier is designed to undertake, the explanation generation method may differ.
Simultaneously the input image is scanned for features that fit the defined features specifications. Pixels of detected features specifications that overlap with explanation pixels are used in calculating the TS.
We discuss below the different parts of our method for calculating TS and we show how following a similar concept of the TS, a suspiciousness score for the image can be calculated using $\{M_{i,1},..., M_{i,J_i}\}$.

% \todo[inline]{Edit method to add, TS threshold, MC threshold, how TS can be used for suspiciousness in frames}

\subsection{Generating Explanations}
The explanations generator in Figure~\ref{fig:method_overview}
aims at identifying which pixels in image $i$ caused the main classifier to output its predictions. Once these pixels are identified an explanation is output where all pixels in image $i$ are blacked out (i.e. set to zero) other than the important pixels which resulted in the main classifier's prediction. Therefore if the main classifier outputs predictions $\{N_{i,1},..., N_{i,Y_i}\}$ the explanations generator should output a set of explanations $\{E_{i,1},..., E_{i,Y_i}\}$ corresponding to each prediction.  

We distinguish between two types of classifiers targeted at different types of object recognition: 1) object classification, where the object classifier only outputs a label of the most probable item in the displayed image. 2) object detection, where the classifier outputs labels and bounding boxes showing the different objects in the image.
For the former, different techniques for generating interpretable explanations for object classification can be used and are available in the literature e.g. GradCam~\cite{Selvaraju2020}, LIME~\cite{Ribeiro2016}, SHAP~\cite{Lundberg2017}, Extremal~\cite{Fong2019}, DeepCover~\cite{Sun2020}. 
All of these methods are developed for generating explanations for classifiers that output only a label. % (i.e. without a bounding box localising the classified object). 
For the latter type of object recognition, the pixels within the bounding box can be considered as the explanation, e.g.\ $N_{i,1}\equiv E_{i,1}$.
%
% sufficient to represent an explanation, as the bounding box encapsulates the pixels resulted in the output prediction i.e. $N(x_i)\equiv E(x_i)$.
%
Throughout the paper we focus our method on object detection, however, the method can be used in the context of object classification too.

\subsection{Features Specifications}
% What are they?
A \textit{specification} generally ``\textit{is a detailed formulation in document form, which provides a definitive description of a system for the purpose of developing or validating the system}"~\cite{ISO24765}\cite{Dhaminda2022a}. 
When defining specifications for features (or features specifications) we aim at unique distinctive properties that make up a classification class. In the rest of this section we discuss how to identify these distinctive features, how to define these features as specifications, and how features specifications can be utilised by machines for monitoring.  

\subsubsection{Identifying Distinctive Features}
 
\begin{itemize}
    \item 
    % \textbf{
    \underline{Method 1, Human logic:} 
    Human labeled data are usually the ground truth used to validate against when it comes to classification. Inspired by that, this approach relies on humans providing a list of distinctive features that has been used in making their personal assessment when labeling data. 
    % In some classification tasks this is very straight forward, e.g. for classifying a person in an image, the distinctive features of person can be the facial landmarks of a human~\cite{Kazemi2014}.
    %
    % \underline{Step1:} 
    \textit{Step~1:} Determine the different classes that the DNN is meant to classify. 
    %
    % \underline{Step2:} 
    \textit{Step~2:} List the distinctive features of each class by analysing through the available dataset. 
    %
    % \underline{Step3}: 
    \textit{Step~3:} Compare between the different distinctive features that were identified for each class and ensure there are no shared distinctive features between classes. In the case where there are shared features, one may want to reflect that in the weighing contribution from that feature in the TS discussed in section~\ref{TS_section}.
    %Identify any shared distinctive features between the different classes. %These shared distinctive features will affect the trustworthiness calculation of which class was detected.   
    
% Provides better ground for justification when solving accountability related issues.
% \begin{itemize}
    \item \underline{Method 2, Explainability analysis:} Explainability analysis is targeted at classification tasks where identifying distinctive features to be used as specifications is not intuitive, especially when slight variations in these features can be critical for the classification output~\cite{Antoran2020} e.g. distinctive features of hand written digits.
    The concept utilises powerful explainability techniques readily available in the literature such as LIME~\cite{Ribeiro2016}, SHAP~\cite{Lundberg2017}, CLUE~\cite{Antoran2020} etc., whereby 
    %
    % \underline{
    \textit{Step~1:} a DNN is trained on the data representing the operational environment (training data), 
    %
    % \underline{
    \textit{Step~2:} the training data is fed into the trained model and an explanation is generated for the different data points, 
    %
    % \underline{
    \textit{Step~3:} an analysis is conducted on the generated explanations using a human to identify which features used as common explanations between data points should be regarded as distinctive.
\end{itemize}

\subsubsection{Defining Features as Specifications}
% \todo[inline]{Revise this section}
The different visual features that collectively make up an object in real life, may be challenging to express rigorously using text specifications solely. Therefore, in defining a feature specification, we comprise the specification of two parts: 1) a specification-text and 2) and a specification-visual. 
The goal of the specification-text is to describe in natural language what subpart of the object is distinctive and requires monitoring. The specification-visual aids with understanding the extent of the specification-text. 
For example, a specification text for monitoring a human palm can be written very simply as ``A person shall have a hand ". This may be interpreted as only the palm of a human, the palm plus the forearm or the whole arm. The specification-visual goal here is to help minimise scope for misinterpretation. 
The specification-visual can be a number of example images to represent the feature, or can also be extended to creating a dataset, following assurance of machine learning for use in autonomous systems (AMLAS) guidelines \cite{hawkins2021guidance} to build a dataset representing the specification-visual.
\subsubsection{Describing Features Specifications in Machine Interpretable form}

Once unique features specifications are identified, they need to be expressed in a form that allows machines to  monitor them.
\begin{itemize}
\item \underline{Method 1, Geometrical analysis:} For some features specifications operating in limited environments one could employ image processing to detect these features using geometrical analysis. 
Whilst geometrical analysis may be fast and avoids the need to gather data for training, they may result in increased levels of false positives and significant changes in the image processing method used when changing operational environments e.g \cite{Lucian2018}. 

\item \underline{Method 2, Deep Learning:} Many of the features that make up a classification in many operational environments are complex, and hence are difficult to represent using geometrical analysis.
Alternatively, one can utilise other DNN models to learn complex features specifications e.g. \cite{Kazemi2014}\cite{Fahn2017}. 

\end{itemize}

\subsection{Trustworthiness Score (TS)}\label{TS_section}
The trustworthiness score for a given prediction $N_{i,y}$ is calculated using equation~\ref{eq:TS}, where $y\in\{1,..,Y_i\}$. 
$R_{i,y,z}$ represents the intersection area between detected feature $F_{i,z}$ and explanation $E_{i,y}$ for prediction $N_{i,y}$. 
A trustworthy prediction is shown by the prediction having overlapping features specifications that make the trustworthiness score surpass a set \textit{trustworthiness score threshold} ($TS_T$). 
Practically using the threshold $TS_T$, a subset of the calculated trustworthiness scores {$TS_{i,1},...,TS_{i,Y_i}$} can be filtered to detect trustworthy predictions.  
%
% having  the detected features specifications overlapping with the prediction. 
%
A prediction and a feature count as overlapping if $R_{i,y,z} \geq R_{lim}$. We use $a_{y,z} \in \{1,0\}$ as a Boolean switch to indicate if this condition is satisfied or not. $\beta_z$ is included in our equation to act as a hyperparameter for weighing the contribution from different types of detected features specifications.

\begin{equation}
\label{eq:TS}
    TS_{i,y} = \sideset{}{}\sum_{z=1}^{Z_i} \beta_z \cdot a_{y,z} \cdot R_{i,y,z}%\frac{\sideset{}{}\sum_{z=1}^{Z_i} \beta_z \cdot a_z \cdot R_{i,z}}{Z_i} 
\end{equation}

\subsection{Suspiciousness Score (SS)}
We also introduce a suspiciousness score calculated on the overall frame level instead on each prediction. 
We refer to a frame as suspicious if it contains false negatives. % from the predictions made by the main classifier.
%
% We also introduce a suspiciousness score using the idea behind the trustworthiness score, where trustworthiness is credited if expected features specifications are identified in a prediction. 
%
% We refer to a frame as suspicious if it contains false negatives from the predictions made by the main classifier. 
%
% Note that the suspiciousness score is calculated on the overall frame level instead on each prediction. 
%
Using the suspiciousness score, suspiciousness is credited to a frame if the frame contains a large portion of the image with predictions below the model confidence threshold i.e. $MC < MC_T$.
We speculate that this gives an indication that the model has high uncertainty in the objects existing in the frame, therefore false negatives in the frame are likely to exist. 
%
%
% The goal is to identify if the overall frame is suspicious, therefore predictions in the frame should not be relied on.
%
% We speculate that the trustworthiness score can be modified to detect the suspiciousness in a frame. A frame is referred to as being suspicious if it contains false negatives from the predictions made by the main classifier. 
%
%
The suspiciousness score (SS) is shown by equation~\ref{eq:SS}. 
In this case the SS loops over predictions $\{M_{i,1},..., M_{i,J_i}\}$ and sums up the area $R_{i,j}$ of the bounding box for each prediction $M_{i,j}$, where $j\in\{1,..,J_i\}$. 
If SS surpasses some selected \textit{suspiciousness score threshold} ($SS_T$) then the frame is suspicious. 
The SS is complimentary to TS, as it tries to identify the existence of false negatives, whilst the TS metric tries to improve the ratio of true positives to false positives.
\begin{equation}
    SS_i = \sideset{}{}\sum_{j=1}^{J_i} R_{i,j}
    \label{eq:SS}
\end{equation}

\subsection{Determining thresholds}
\begin{figure}[h]
    \centering
    \begin{subfigure}{0.35\textwidth}
        \includegraphics[width=1\textwidth]{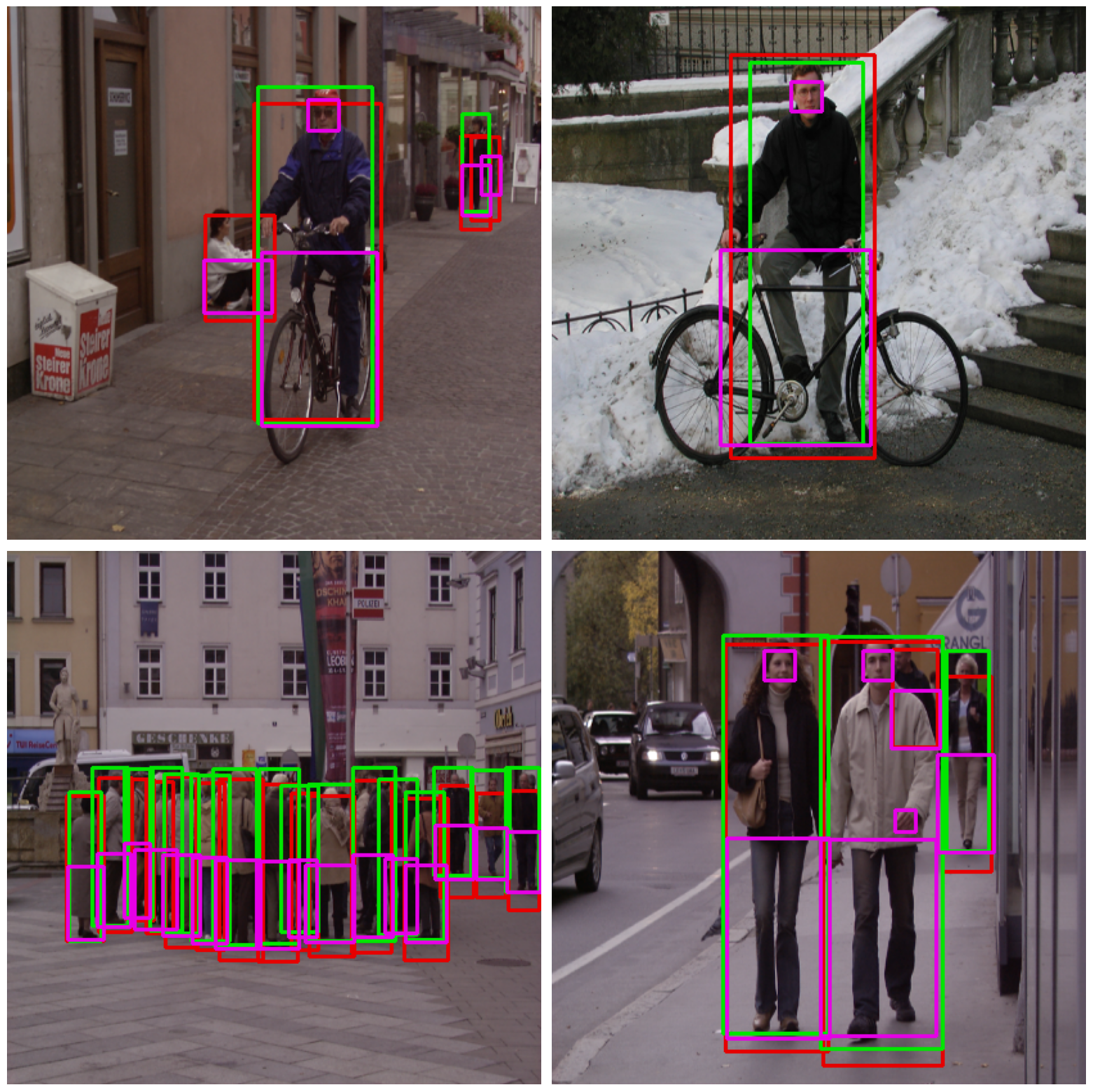}
        \caption{INRIA Person}
    \end{subfigure}
    \begin{subfigure}{0.35\textwidth}
        \includegraphics[width=1\textwidth]{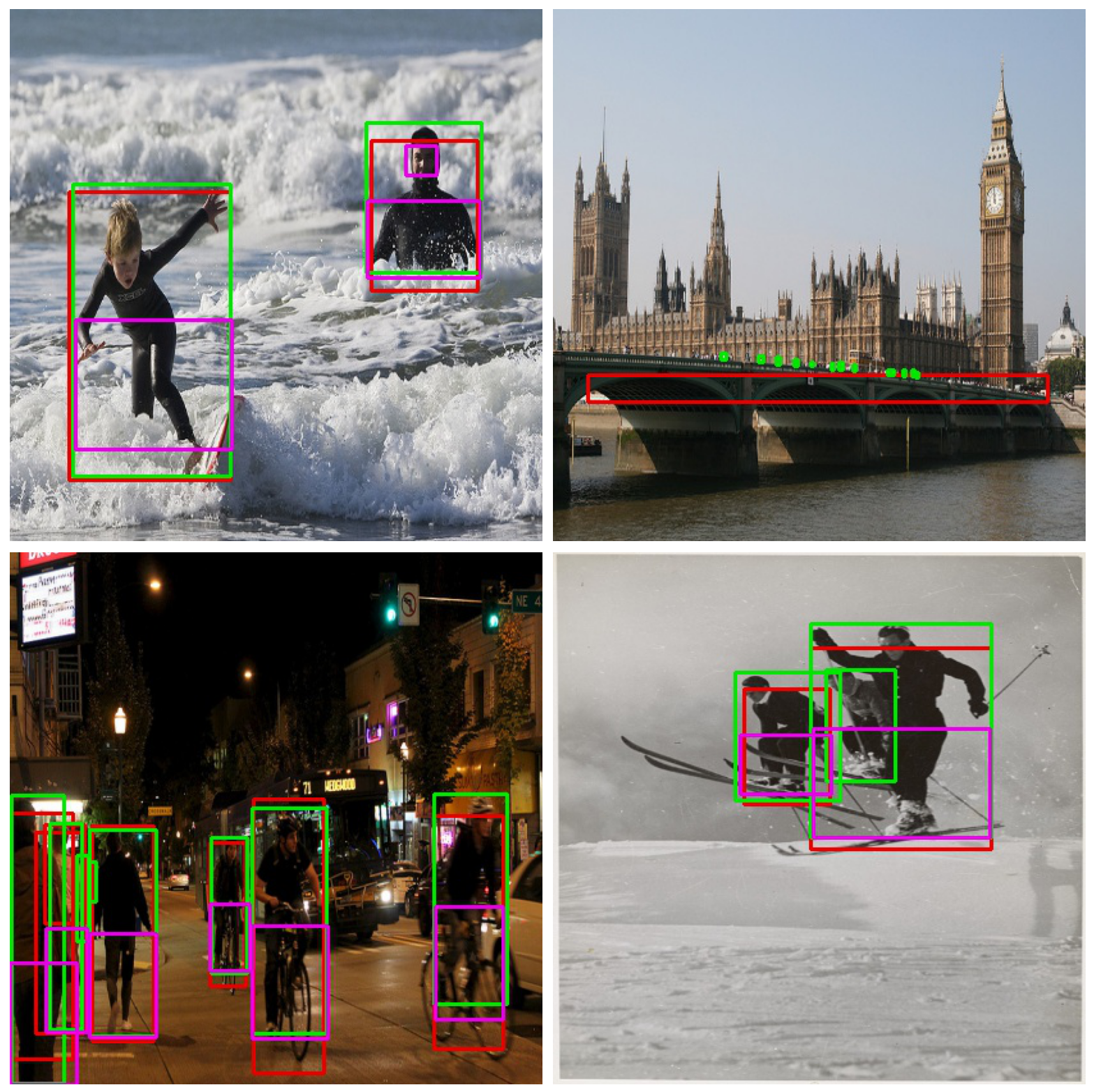}
        \caption{COCO}
    \end{subfigure}
    \caption{Images from INRIA and COCO datasets showing main classifier predictions (red), ground truth annotations (green) and detected features specifications (magenta).}
    \label{summary_pics}
\end{figure}

There are four thresholds involved in our method: $MC_T$, $R_{lim}$, $TS_T$ and $SS_T$.
Determining threshold values may depend on the application domain, specifications, performance requirements, or regulatory aspects. 
We do not focus on methods for determining these thresholds, however in our experimentation we evaluate our method on the whole range of $MC_T$ (0 to 100), whilst $TS_T$ and $SS_T$ are tuned to maximise F1-Score at the different $MC_T$ values.
%
% Determining $R_{lim}$ depends on the application and type of features specification being monitored. 
% Formalising a method for determining $R_{lim}$ can be scope for future work. 
In our experimentation we have set $R_{lim}$ to $70\%$, as through visual assessment of the outputted bounding boxes (see Figure~\ref{summary_pics}) we deem this to be sufficient to confirm an overlap exists.

% ========================
% Experimentation Setup
% ========================
\section{Experimentation Setup} \label{sec:experiment}
We used the application of person detection using DNN to advocate our method in both identifying trustworthy predictions and detecting suspicious frames. 
The effectiveness of trustworthiness detection in predictions using the trustworthiness score (TS) method is evaluated against model confidence (MC) as our baseline. 
The baseline for assessing the effectiveness of the suspiciousness score (SS) in detecting suspicious frames is also evaluated against MC.
For SS case, the baseline considers a frame to be suspicious if the main classifier outputs predictions with model confidence below the $MC_T$. 
In the rest of this section we discuss further details for our experimentation setup. Code is available at this repository: \url{https://github.com/Abanoub-G/TrustworthinessScore}

\subsubsection{Main Classifier}
For any given application, the \textit{main classifier} is the DNN responsible for detecting the object of interest, in this case study the object of interest is a person. 
A YOLO~\cite{Redmon2015} object detector pre-trained on the COCO dataset~\cite{COCO_dataset} was chosen to act as the main classifier in our case study. The choice of using YOLO as the main classifier was based on its growing utilisation and popularity in the machine learning community. In addition, to the simplicity provided in its integration via PyTorch and the documentation availability through GitHub~\cite{Jocher_YOLOv5_by_Ultralytics_2020}. 
The YOLO object detector outputs a label, bounding box and model confidence for the detected objects. The bounding box can conveniently serve as the explanation to the prediction, whilst the model confidence provides the baseline to evaluate the effectiveness of the trustworthiness score.

\subsubsection{Features specifications}
In the task of detecting a person we identify (using human logic) three main distinctive features. We list them as features specifications below\footnote{The method used in writing down the specifications is based on NASA guidelines for writing requirements and specifications.~\cite{NASA_SystemsEngineeringHandbook}.}:
\begin{itemize}
    \item A person detection in an image shall include human facial landmarks.
    \item A person detection in an image shall include a palm.
    \item A person detection in an image shall include human legs.
\end{itemize}

For all three listed features specifications we used object detectors to describe the features in machine interpretable form for monitoring. 
As observed we only provided the specification-text part of a feature specification and eliminated the usage of a specification-visual. 
This is becuase we used existing pretrained object detectors as monitors for our features specifications, allowing us to overcome the need to build datasets for the specification-visual.  

Pre-trained models for detecting faces and hands were used to monitor for the first two specifications. 
For detecting faces a pre-trained model based on Dlib was used~\cite{king2015}.
For the detection of palms MediaPipe-Hands by google research was used~\cite{mediapipe}.
For the the third specification of legs detection, a pre-trained model for detecting legs was not found, therefore, we trained our own custom legs object detector using the YOLO5s~\cite{Jocher_YOLOv5_by_Ultralytics_2020} architecture on a synthesised legs dataset. The legs dataset was synthesised by taking the lower half of the person labels in the COCO dataset and labeling them as legs.
\subsubsection{Datasets}
Two datasets were used in our evaluation: 1) INRIA Person dataset~\cite{Dalal2005} (size: 902 images) and 2) COCO dataset~\cite{COCO_dataset} (size: 328,000 images). 
The INRIA dataset has served as a lightweight dataset for development, whilst the COCO dataset is used to ensure the method scales.% to larger datasets.

\subsubsection{Hyperparameters selection}
There are two hyperparameters associated with the trustworthiness score (TS): $\beta$ and $TS_T$. For all experiments, we set $\beta = 1$ for all features specifications. The $TS_T$ (or the $SS_T$ in the case of detecting suspiciousness) is varied to maximise the F1-Score of the TS method approved predictions.

% ========================
% Results and Discussion
% ========================
\section{Results and Discussion} \label{sec:results}
To evaluate our method we analyse against ground truth the True Positives (TP), False Positives (FP) and False Negatives (FN) for trustworthy predictions based on 1) the model confidence only then 2) the trustworthiness score.  
The metrics often used to analyse TP, FP and FN are Precision, Recall and F1 score. Precision is calculated as $\frac{TP}{TP+FP}$  which is meant to convey how many of the predictions are accurate out of all predictions. Recall on the other hand, expresses the rate of the classifier getting correct predictions. This is calculated as $\frac{TP}{TP+FN}$. Often as one tries to increase Precision, Recall tends to decrease as a result. Ideally one aims to maximise both Precision and Recall. The F1 score is used to express the maximisation of both precision and recall simultaneously. This is given by $\frac{2\cdot Precision\cdot Recall}{Precision + Recall}$.
The aim of our approach is to improve Precision whilst keeping the F1 Score approximately the same or improving it. 
Therefore, we mainly focus on Precision improvements but we also monitor F1 score. 

\begin{figure*}[h]
    \centering
    \begin{subfigure}{0.42\textwidth}
        \includegraphics[width=1\textwidth]{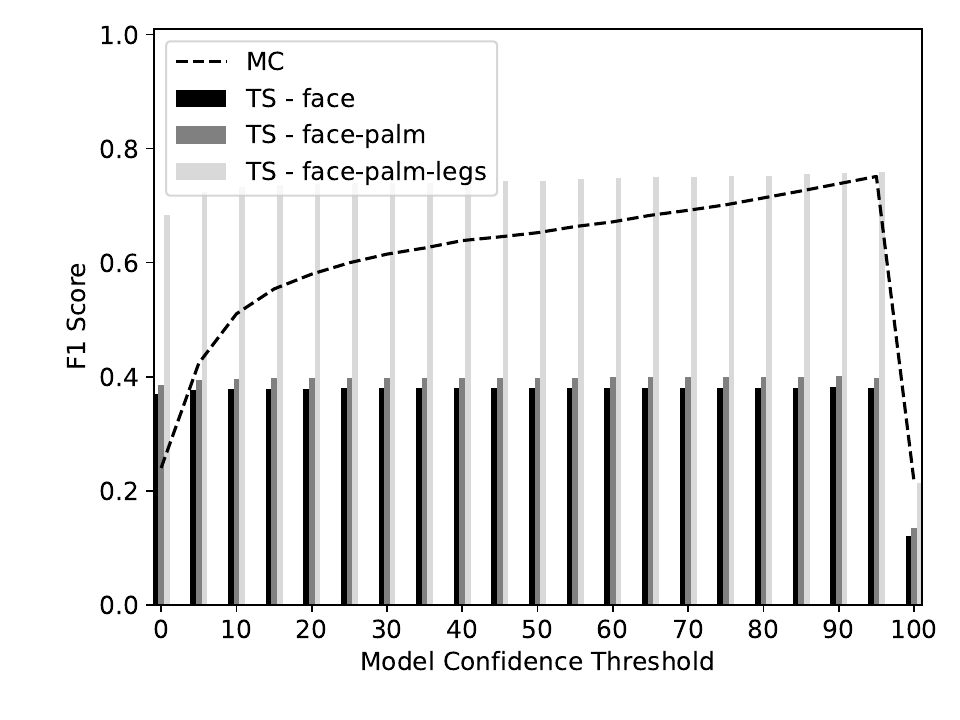}
        \caption{INRIA Person}
    \end{subfigure}
    \begin{subfigure}{0.42\textwidth}
        \includegraphics[width=1\textwidth]{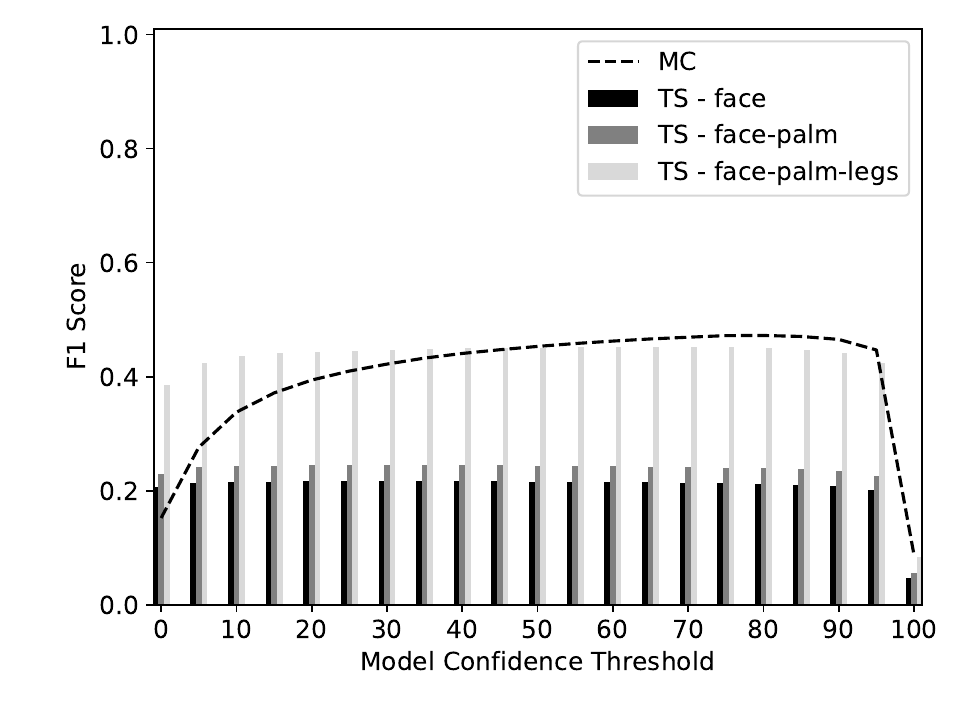}
        \caption{COCO}
    \end{subfigure}

    \begin{subfigure}{0.42\textwidth}
        \includegraphics[width=1\textwidth]{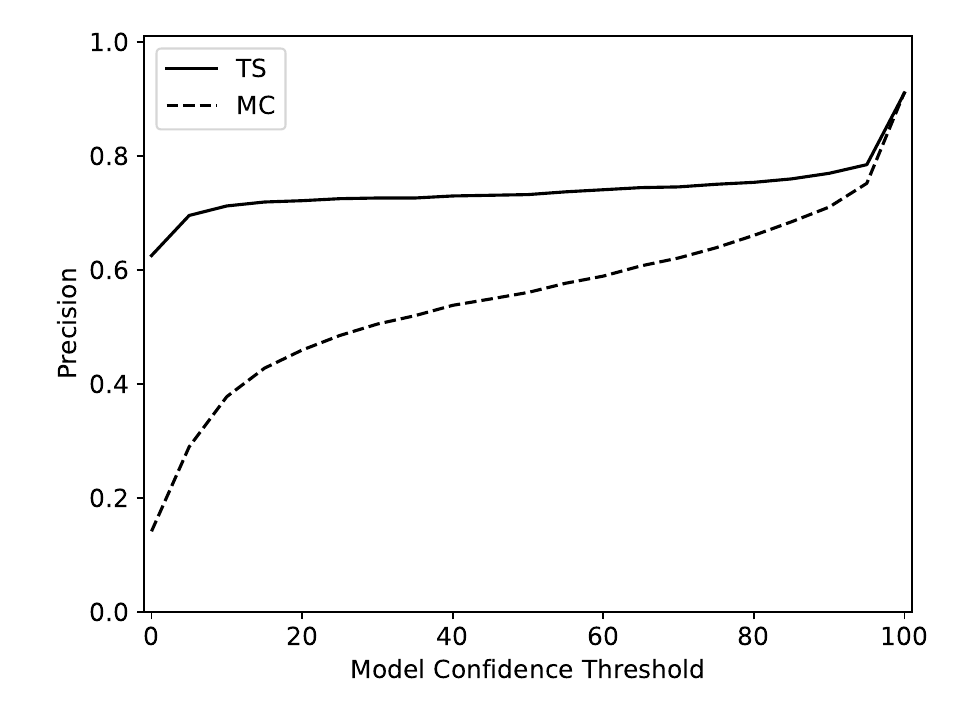}
        \caption{INRIA Person}
    \end{subfigure}
    \begin{subfigure}{0.42\textwidth}
        \includegraphics[width=1\textwidth]{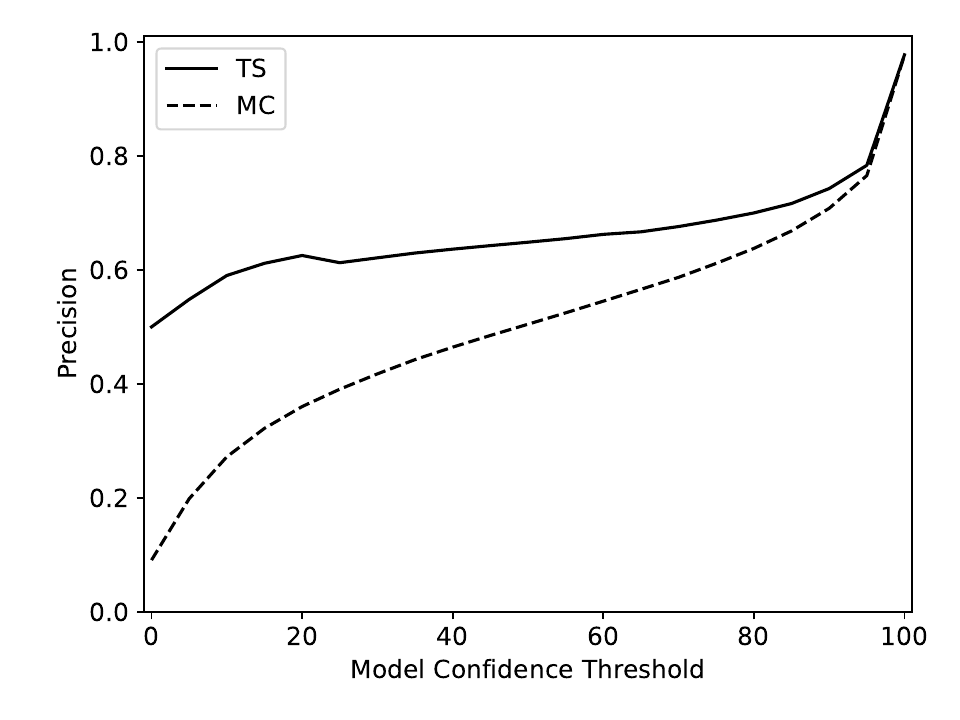}
        \caption{COCO}
    \end{subfigure}
    \caption{Detect Trustworthiness}
    \label{detect_trustworthiness}
\end{figure*}

\begin{figure*}[h]
    \centering
    \begin{subfigure}{0.42\textwidth}
        \includegraphics[width=1\textwidth]{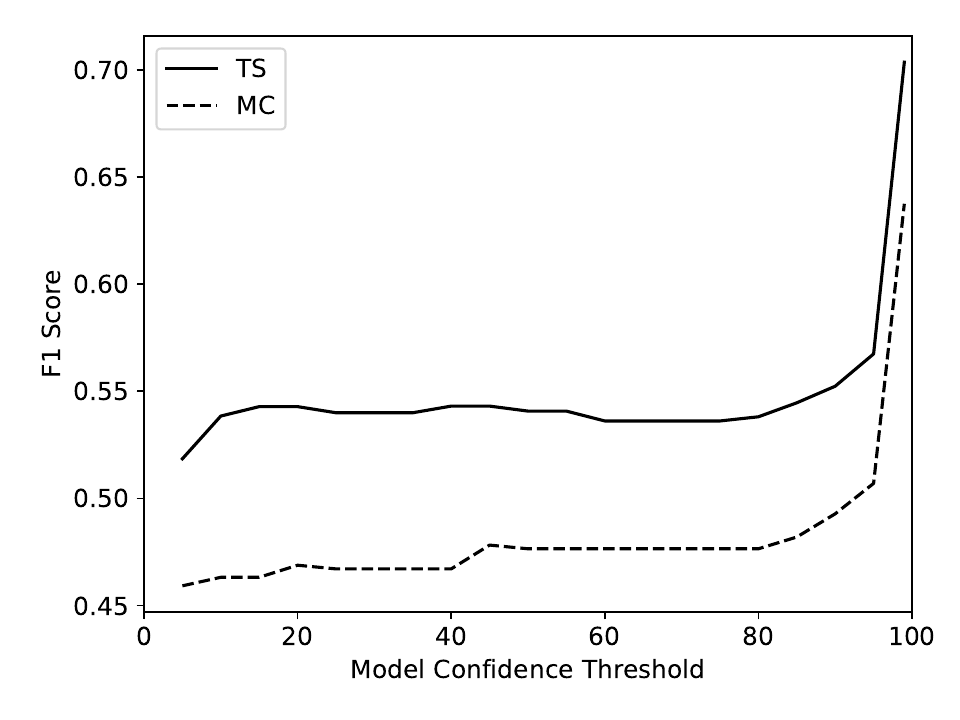}
        \caption{INRIA Person}
    \end{subfigure}
    \begin{subfigure}{0.42\textwidth}
        \includegraphics[width=1\textwidth]{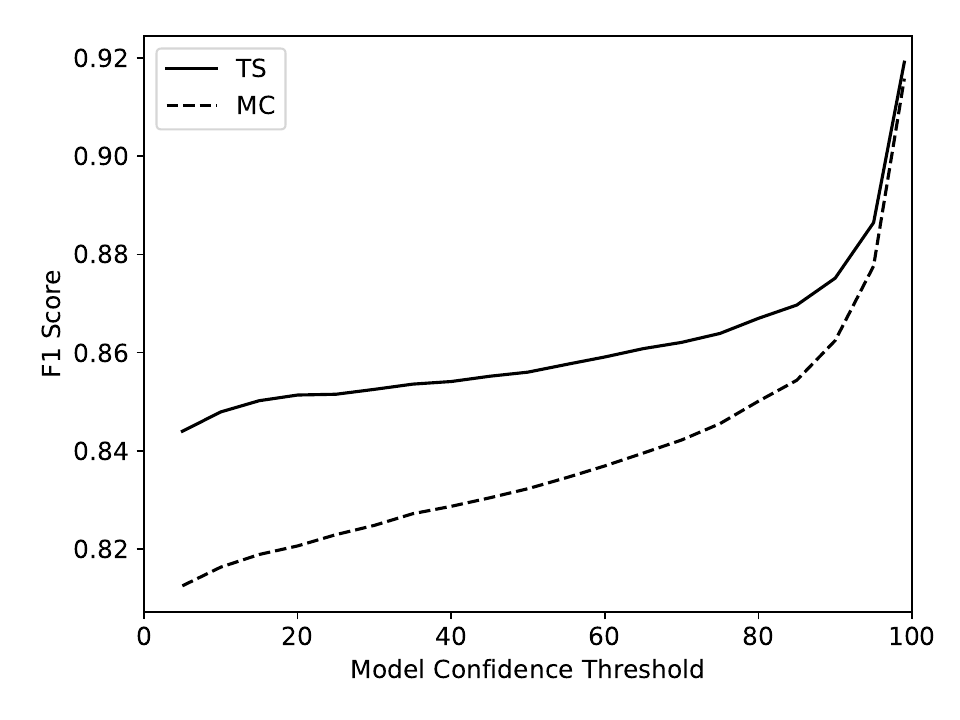}
        \caption{COCO}
    \end{subfigure}

    \begin{subfigure}{0.42\textwidth}
        \includegraphics[width=1\textwidth]{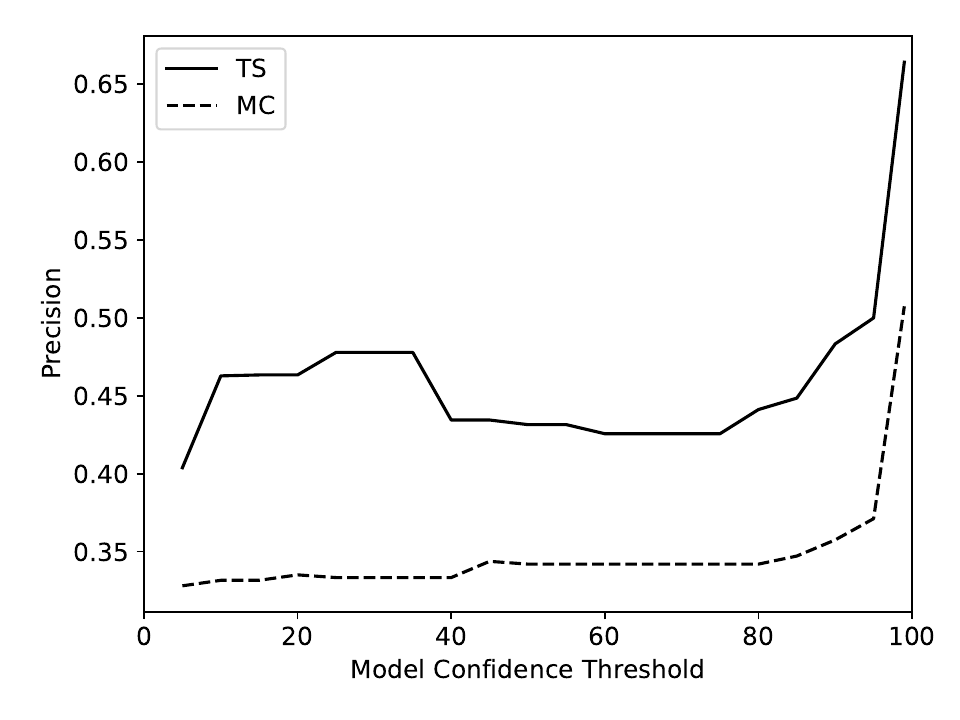}
        \caption{INRIA Person}
    \end{subfigure}
    \begin{subfigure}{0.42\textwidth}
        \includegraphics[width=1\textwidth]{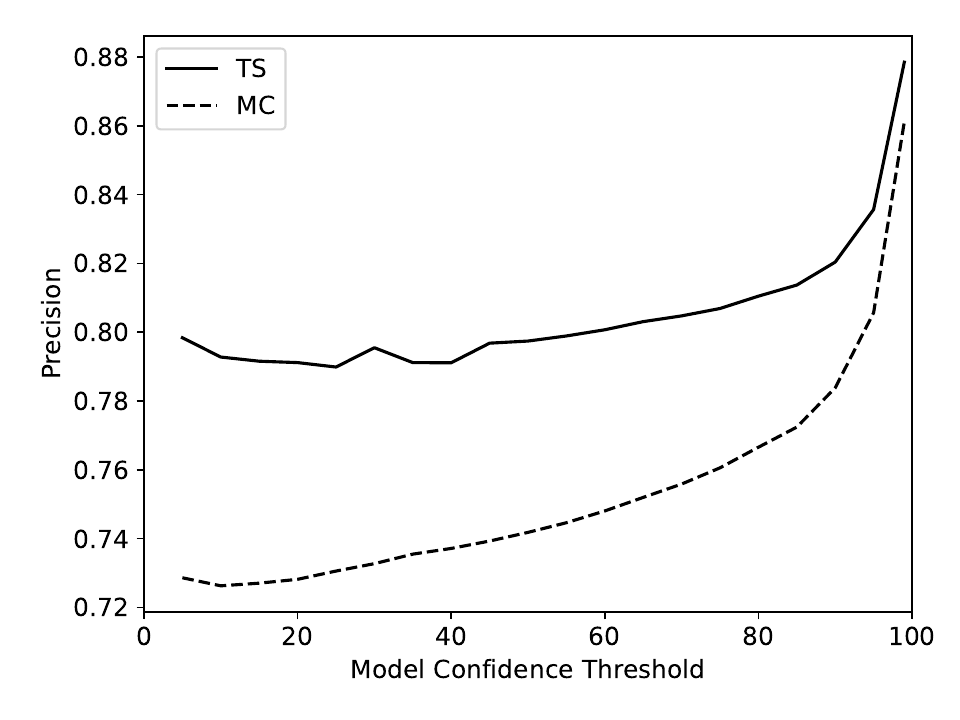}
        \caption{COCO}
    \end{subfigure}
    
    \caption{Detect Suspiciousness}
    \label{detect_suspiciousness}
\end{figure*}
\subsection{Features specifications sufficiency}
Under representation of objects using features specifications can lead to F1 Score based on TS to be less than on MC. Therefore, the first criterion we assess is the sufficiency of features used to represent the object being classified.
Figure~\ref{detect_trustworthiness} (a) to (b) shows the F1 score of the main classifier for INRIA and COCO datasets. The F1 score when relying on model confidence alone to approve of trustworthy predictions is shown by the dotted line. The bars represent F1 score when relying on TS to approve of predictions made by the main classifier. The bars are shown for three TS setups, in each setup a feature specification is added incrementally. 
The results are shown for different accuracies of the main classifier. 
As can be seen for both datasets in Figure~\ref{detect_trustworthiness}, the F1 score for our TS method gives poorer performance than MC when using only face as a feature specification, with a slight increase when adding hand feature specification, and then surpasses the MC performance when adding in legs feature specification. 
This shows how sufficient representation of an object using feature specifications is crucial for effectiveness of our approach.

\subsection{Detect Trustworthiness}
Figure~\ref{detect_trustworthiness} (c) to (d), shows how the performance based on precision changes with the accuracy of the main classifier for both the INRIA and the COCO datasets. It can be noticed that the trustworthiness score consistently performs better than the model confidence alone, with the amount of improvement decreasing as the accuracy of the main classifier increases. 
It can be observed using Figures ~\ref{detect_trustworthiness} (a) to (d), that the trustworthiness score approach significantly improves the precision, whilst also improving or maintaining the F1 score to a level equivalent to that provided by the model confidence alone.  
Observing the precision plots in Figure ~\ref{detect_trustworthiness} our method provides an improvement of approximately 40\% for low model confidence thresholds and diminishes to zero for high model confidence thresholds, averaging to $\approx 20$\% improvement when averaged across different model confidence thresholds.

\subsection{Detect Suspiciousness} 
Similar to results shown for detect trustworthiness, Figure~\ref{detect_suspiciousness} (a) to (d) show cases the effectiveness of the SS metric to calculate the suspiciousness in a frame. 
%This is compared against the model confidence as our baseline. 
As can be seen the SS provides a significant improvement compared to relying on MC only. Similar to our observation in detect trustworthiness, this effectiveness diminishes as the model confidence threshold of main classifier increases.
Examining the precision plots in Figure ~\ref{detect_suspiciousness} our method provides an improvement of approximately 10\% for low model confidence thresholds and diminishes to zero for high model confidence thresholds, averaging to approximately 5\% improvement when averaged across different model confidence thresholds.

\subsection{Further Remarks and Limitations}
The overall trustworthiness of artificial intelligence is challenging to define using a single-score metric. 
There are multiple factors that affect trustworthiness, many of which depend on the application context~\cite{Wing2021,9142644,thiebes2021trustworthy}. 
Therefore, overall trustworthiness requires a multivariate scoring approach. Spider web diagrams (or radar charts) may present an optimal way for visualising overall trustworthiness and the trade-offs that can be made in different aspects of trustworthiness. 
In a spider web diagram each variable is represented by a separate axis, each representing the scale of the variable being measured.
Data points calculated for each variable affecting trustworthiness are plotted on the axes and connected to form a polygon.
The shape and area of this polygon represent overall trustworthiness in a more tangible way than what can be communicated  using a single-score metric.
Our TS metric uses an approach of explainability to enhance the performance of DNN. It can, hence, also be included in overall trustworthiness spider web diagrams as a metric on the axis of explainability.

The approach of using features specifications for calculating TS can be viewed in the context of assertions monitoring. 
Assertions are statements in programming languages that check if a certain condition is true, and if not, they raise an error.
They are typically used to ensure that the code behaves as expected e.g. checking behavioural compliance of self-driving vehicles with highway code rules~\cite{harper2021safety}.
The features specifications we monitor can be seen as assertions used to indicate lack of trustworthiness when monitored features are absent.
% an error when 
% being monitored can be would represent assertions that are expected to overlap with the main classifier predicitons the  being monitored to identify when build confidence Very similar to an approach introduced by Harper et.al.~\cite{harper2021safety} of using him converting 

Whilst the TS presents a promising approach for improving detection of trustworthy predictions and suspicions frames during operation, the method may still suffer from out of distribution classification issues, especially that features specifications detection relies on other DNN.
Therefore, methods for detection of out of distribution environments (e.g. dissimilarity measures~\cite{Hond2020}).) are very complimentary to our method.

% ========================
% Conclusions
% ========================
\section{Conclusions} \label{sec:conclusions} 
We have introduced a method for quantifying and calculating trustworthiness score (TS) in a classifier's predictions by monitoring for features specifications.
We found that the number and type of monitored features specifications used in calculating TS is critical to the effectiveness of the approach.
We have also explored if a modification of TS can be used to help with the detection of suspicious frames (i.e. frames containing false negatives) via calculating a suspiciousness score. 
Overall, our method showed an improvement compared to using model confidence of a classifier in detecting trustworthy predictions and suspicious frames. 
Our method provides an averaged improvement of 20\% and 5\% for detecting trustworthiness and suspiciousness respectively. 

\section*{Acknowledgments}
This research is part of an iCASE PhD funded by EPSRC and Thales UK. 
Kerstin Eder was supported in part by the UKRI Trustworthy Autonomous Systems Node (grant number EP/V026518/1).
Thanks to Edwin Simpson from University of Bristol for helping with reviewing.

% \balance

% ***************************************************
%  Bib
% ***************************************************
\printbibliography

% ***************************************************
%  Appendix
% ***************************************************

\end{document}